\documentclass[conference]{IEEEtran}

\usepackage[pdftex]{graphicx}
\usepackage{multirow}
\usepackage{verbatim}
\usepackage{booktabs,mathptmx,siunitx}
\graphicspath{{figures/}}
\DeclareGraphicsExtensions{.pdf,.jpeg,.png}
\usepackage[ruled]{algorithm2e}
\SetAlFnt{\small}

\usepackage{color}
\ifCLASSINFOpdf
\else
\fi
\usepackage[caption=false,font=footnotesize]{subfig}
\begin{document}
\title{{\em dMath}: A Scalable Linear Algebra and Math Library for Heterogeneous
GP-GPU Architectures
}

\author{\IEEEauthorblockN{Steven Eliuk, %\IEEEauthorrefmark{1},
Cameron Upright%\IEEEauthorrefmark{1} and Anthony Skjellum\IEEEauthorrefmark{2}}
}
\IEEEauthorblockA{%\IEEEauthorrefmark{1}
Samsung Electronics\\ 
Computing Science Innovation Center, SRA-SV\\
665 Clyde Avenue\\
Mountain View, CA 94043\\
Email: \{s.eliuk,c.upright\}@samsung.com}
\and
\IEEEauthorblockN{Anthony Skjellum}
\IEEEauthorblockA{
RunTime Computing Solutions, LLC\\
1500 1st Avenue North, Unit \#19\\
Birmingham, AL 35203\\
Email: tony@runtimecomputing.com}}


%Telephone: (800) 555--1212, Fax: (888) 555--1212}


\maketitle

%%Introduction/Motivations
%Architecture of the Underlying System (Hetereogeneous CPU-GPU-IB)
%DMath Library Strategy
%a) Client-Server model -- design for use by programs that do not %want to do explicit
%b) Encapsulation and abstraction
%c) Managing persistent data in the GPU
%d) Data-reorganization and data service model
%e) Data distribution independent algorithms
%f) examples of performance and speed up
%g) Heterogeneity issues
%Application Implications
%i) Ease of use in programs that do analytics and need algorithmic speed up
%ii) Maturity of the underlying COTS systems (MPI, hardware, drivers, ...)
%Future Work
%Conclusions.

%%%%%%%%%%%%%%%%%%%%%%
%%%-------ABSTRACT-----%%
%%%%%%%%%%%%%%%%%%%%%%
\begin{abstract}
%\boldmath
% abstract cant have citations.....
A new scalable parallel math library, {\em dMath}, is presented in
this paper that demonstrates leading scaling when using intranode, or internode, hybrid-parallelism
\nocite{Krizhevsky14} for deep-learning.  {\em dMath} provides easy-to-use distributed base primitives and a
variety of domain-specific algorithms.  These include matrix multiplication,
convolutions, and others allowing for rapid development of highly
scalable applications, including Deep Neural Networks (DNN), whereas previously one was restricted
to libraries that provided effective primitives for only a single GPU, like Nvidia's cublas \& cudnn
\nocite{DBLP:journals/corr/ChetlurWVCTCS14} 
or DNN primitives from Nervana's \nocite{nervana_neon}neon framework.

Development of HPC software is difficult, labor-intensive work,
requiring a unique skill set.  {\em dMath} allows a wide range of
developers to utilize parallel and distributed hardware easily.  One
contribution of this approach is that data is stored persistently on
the GPU hardware, avoiding costly transfers between host and device.
Advanced memory management techniques are utilized, including caching
of transferred data and memory reuse through pooling.  A key
contribution of {\em dMath} is that it delivers performance, portability,
and productivity to its specific domain of support.  It enables
algorithm and application programmers to quickly solve problems without managing the significant complexity
associated with multi-level parallelism.  {\em dMath} can use intranode GPU-Direct Remote Direct Memory Access (GDR),
developed in collaboration with the OpenMPI and MVAPICH groups that
has shown to decrease latency and increase bandwidth when compared to
previous techniques.  Efficient inter-GPU communication is crucial to
achieving greater net performance and supporting effective use of the
cost-effective, GPU-dense COTS architecture adopted.
{\em dMath}'s caching approach addresses
one of the key drawbacks of GPUs, which is to keep data sets cached
and to avoid overheads of the CPU-GPU memory interface wherever
possible.

\end{abstract}


% \fi
\IEEEpeerreviewmaketitle

%%%%%%%%%%%%%%%%%%%%%%
%%%-------INTRO-----%%
%%%%%%%%%%%%%%%%%%%%%%
\section{Introduction}
Machine learning algorithms leverage traditional scientific
computing--- correlations, convolutions, FFTs, matrix and tensor
multiplication, and combinations thereof.  Thus, central to the solution
of key machine learning algorithms such as stochastic gradient descent
\cite{SGD-wikipedia} is the need both for scalable architectures and
algorithmic libraries that implement these kernels efficiently.
Strong (Amdahl's law) scaling\footnote{That is, minimum time to solution is the goal.}is 
the appropriate performance metric \cite{amdahls-law}, because
problems are not field-based, and do not scale to support weak scaling
(Gustasfson-Barsis or scaled speedup) \cite{gustafson-barsis}.

The emergence of COTS x86-64 multicore servers as the hardware
platform together with successive generations of faster and faster PCI buses
and fabrics complete the overall picture needed to create COTS-based
heterogeneous systems---from the hardware perspective---that are tuned for fast machine learning.
However, a suitable programming model that supports both high
performance in single machines (utilizes the GPUs for data-parallel
and task parallel concurrency) and exploits the scale-out feature of the
systems (medium-grain, data parallel concurrency achieved with message passing) is needed.  Users
will rarely be willing to write data-parallel programs in MPI that couple with
local vectorized/GPU-enabled math libraries or with MPI-based libraries.  Instead, they seek ease of
use.  Furthermore they will often want to alternate between data parallelism and task
parallelism, such as in-memory analytics ({\em e.g.}, Spark
\cite{spark}). Furthermore, higher productivity, and abstraction of the algorithms
and details of dealing with the various sources of performance are needed,
and MPI and scalable libraries of the traditional sort are insufficient.

The vast majority of users wish to solve problems involving
machine learning and data analytics in a timely, efficient manner.  Most often they do not
want to become parallel programming experts
in order to exploit the performance of GP-GPU-enabled high-speed
clusters.  Their primary goal is algorithmic design and evaluation, and utilization in applications.
In all cases, time to solution of fixed sized problems is the chief concern for such users.

To address these opportunities and challenges, 
this paper presents {\em dMath}, a new scalable distributed math library.  {\em dMath} provides key linear algebra operations, convolutions and other fundamental algorithms useful in the implementation of Distributed Neural Networks (DNNs).  Other examples, such as cuDNN from Nvidia~\cite{DBLP:journals/corr/ChetlurWVCTCS14}, provide primitives for Deep Learning (DL) on a single GPU.  In contrast, {\em dMath} provides primitives for distributed DL.  We will identify several key components that have helped in the rapid development and scalability of dMath.

Fundamental features of this library include:
a) support for persistent storage of operands in the GPU's device memory to avoid the CPU-GPU bottleneck;
b) the ability to exploit multiple paths for data transfer;
c) a novel data management service that allows caching of objects shared through data parallel operations for later reuse;
d) services for data reorganization to support optimization of operations in series; 
e) an object-oriented design that abstracts multi-GPU, multi-server computing from the end-user; and 
f) an effective master-worker model to allow users (including those performing in-memory analytics and other task parallel operations) to utilize dMath without requiring detailed knowledge of CUDA, MPI, or data reorganization.  Performance and efficiency of GPU computations (strong scalability) are emphasized.
Flexible data layouts of matrix objects are supported as well. This library also demonstrates the ability to use single, double, and half-precision floating point in support of key parallel algorithms.
Fast type conversion, lossy compression, sharing of data at lower precision, and mixing heterogeneous half and float precision operations are emerging features.  Finally, fault tolerant aspects of this library and applications built thereon are discussed.

The main innovative claims of dMath are as follows:
\begin{itemize}
\setlength\itemsep{0em}
\item dMath looks and feels like a regular math library.
\item dMath provides efficient primitives for distributed DL, and will be demonstrated via \textit{Expresso} -- Caffe powered by this distributed library.
\item Abstract matrix and vector classes are used as the basis for specialized versions with precision, layout and computational targets (CPU, GPU, Distributed).
\item dMath allows user code to be written without direct knowledge of its inner workings, or focus on specifics of the data-parallelism (all levels/kinds).
\item A client-server model is supported, where the user's main thread drives backend parallel computations; the data remains resident in the MPI processes and, specifically, cached in GPU memory.
\item Within concrete implementations of the distributed matrix and vector types,
dMath dispatches work to the worker nodes that collaborate via an MPI communicator.
% we should probably avoid 'slave'... not PC.
\item Eases data reorganization of concurrent objects.
\end{itemize}
A further contribution of dMath is that it achieves leading scaling when using intranode, or internode, hybrid-parallelism
\cite{Krizhevsky14} (See section~\ref{sec:examples}). This also means as model sizes grow one is not restricted to the memory size of a single GPU, like that of data-parallel techniques, but the aggregate of all GPUs chosen to be utilized.

dMath features a data reorganization and replication service that allows for
 reshaping matrices (providing a simple copy mechanism for changing
 concurrency for different stages of operations, and remaps
 between parallel data layouts).  A key application of this feature
 is the optimization of a DNN pipeline.  For instance, the matrix-matrix multiplication (GEMM) level is fastest on a relatively smaller number of GPUs.
 This contrasts with the convolutions which require little to no communication, and scale easily.
 Therefore, the ability to move from one data layout and concurrency to another is key to
 achieving good overall performance (rather than making compromises between stages to keep data statically laid out).

With large amounts of GPU memory
 and extensive computational functionality, data and computation
 remains persistent on the GPUs.  Avoiding transfers between host and
 device avoid crippling scalability.  dMath thereby addresses
 one of the key challenge of GPUs, which is the difficulty of keeping data sets cached
 and avoiding the overhead of the CPU-GPU memory interface wherever
 possible.  Furthermore, a key feature of the memory management within nodes offered by dMath is
pooling of unused GPU memory that avoids the costly CUDA allocation and registration with the IB driver.

Another notable feature of the system is the ability to ``keep what
you've seen.''  Because the data management layer has semantic
understanding of matrices and vectors, as certain algorithms (such as
a Cyclic GEMM) progress, portions of the parallel matrix can be retained in a cache
(within each MPI process).  This allows for reduced communication in subsequent
steps, such as in the back-propogation stage of DNN
training.  The fact that the systems can store the output data unscalably leads
to better overall performance, and points to the efficacy of the GPU
architecture for certain classes of strong scaling problems.  For problems
where computations are memory bound, replication is disabled.

Last, we exploit GPU-enabled MPI to enhance performance and exploit 
non-blocking MPI operations to address overlapping of communication
and computation through double buffering, where appropriate.  These steps
are often difficult for application programmers to implement; solving
these problems in dMath simplifies development. 

\nocite{GDR-reference1,GDR-reference2}
%By working with OpenMPI and MVAPICH2, we are able to take advantage of
% define GDR terminology first use... or use GDDR if decided that's more standard
%GPU-Direct RDMA (GDR) \cite{GDR-reference1,GDR-reference2} for certain operations.  
%The maturity and strengths and weaknesses
%of GPU-enabled MPI are discussed as well, including the need for
%additional operations (near future work).  
%We also explore the need
%for means to extend MPI to allow for non-blocking operations that have
%pre-, or post-, GPU kernels applied, such as when we convert between
%precisions in conjunction with network transfer to conserve bandwidth
%for algorithmic steps that can tolerate such lower precision.
%\textit{\textbf{Eliuk *this needs to be further discussed  in the results section or removed}}

The remainder of this paper is organized as follows.
Section~\ref{sec:background} provides background on the motivations
for the creation of an heterogeneous, multinode architecture suitable
for high performance algorithms, machine learning, and analytics based
on COTS components.  It also addresses the motivations for the creation
of the dMath library and programming model.  
The system architecture underlying dMath is described in
Section~\ref{sec:HW_Arch_Considerations}.
A detailed 
explanation and motivation for the dMath architecture is given in
Section~\ref{sec:dmath-architecture}.  
Section~\ref{sec:related-work} highlights similarities and
differences of dMath to related work.
Section~\ref{sec:HW_Arch_Considerations} describes
that COTS architecture, including the synergistic combination of
latest generation NVIDIA GP-GPUs, 100Gbit/s Mellanox InfiniBand, and
high performance PCI-Gen3 Xeon Haswell servers.  
%This follows with an
%explanation and motivation for the dMath architecture in
%Section~\ref{sec:dmath-architecture}.  
Examples of performance and
speedup are then given in Section~\ref{sec:examples}.  In
Section~\ref{sec:heterogeneity}, we discuss the issues that
heterogeneity introduces into the system including non-uniform
performance of the multiple network connections of the architecture.
We also address the maturity of the COTS components including MPI
middleware.  %Comparisons to well-known approaches to data-parallel
%libraries based on MPI and used in GPUs are covered in
 Future Work ({\em cf.},
Section~\ref{sec:future-work}) is covered next, including how the
architecture and dMath are evolving to exploit new features of GPUs
and systems, followed by conclusions in Section~\ref{sec:conclusion}.

\section{Background}
\label{sec:background}
High-speed machine learning is becoming one of the most important areas of high
performance computing~(HPC), in the commercial sphere, in acedemia and elsewhere.
Cost-effective, scalable machine
learning is a crucial aspect in the solution of significant problems.
Time to solution and the potential for online
applications include cloud-based and enterprise-based services such
as mobile applications.  However, off-the-shelf commodity clusters
need to be optimized both in software and hardware dimensions 
in order to support optimized data-parallel machine learning algorithms and task-parallel, in-memory
analytics.

Since the early 1980's ({\em e.g.}, \cite{Seitz:1985:CC:2465.2467}), systems based
on communicating sequential processes \cite{Hoare:1978:CSP:359576.359585}, data parallelism \cite{data-parallelism}, and shared-nothing message passing have proven
the most scalable for massive scale-out computations and large
memories.  These architectures have evolved into cluster computers with high speed COTS
interconnects such as Myrinet~\cite{Boden:1995:MGL:623261.623898}, and now predominantly,
InfiniBand~\cite{Hamada-infinibandtrade}.  The most effective source of raw high
performance floating point performance has emerged in the last decade
as long-vector accelerators typified by Nvidia and AMD General-Purpose
Graphical Processing Units (GP-GPUs).  Multicore CPUs have not kept up with GPU hardware;
far less attention to floating point has been given, despite the emergence of short-vector instruction sets
({\em e.g.,} AVX2 \cite{avx2}) on x86-64 processors.

In the area of mathematical libraries, much work has been done for
multiprocessor, multicore, and multicomputer architectures, typified by
ScalaPack \cite{Dongarra:1997:SUG:265932}, Plapack \cite{conf/ppsc/AlpatovBEGMOGW97}, LAPAck
\cite{Anderson:1990:LPL:110382.110385}, BLAS \cite{Boisvert:2002:PSI}, and more recently BLIS \cite{VanZee:2015:BFR:2786970.2764454}.
These libraries comprise de facto API standards.  In
the areas of vector-accelerated math libraries, Nvidia provides high
performance FFTs \cite{cufft} and matrix algebra cuBLAS \cite{cublas} for GP-GPUs, in addition to
third-party offerings.

%%%%%%%%%%%%%%%%%%%%%%
%%%-------BODY SECTION 2-----%%
%%%%%%%%%%%%%%%%%%%%%%
\section{{\em dMath} Architecture}
%a) Client-Server model -- design for use by programs that do not want to do explicit
%\subsection{Sequential Server-Data Parallel Client Model}
\label{sec:dmath-architecture}
%{\em Note: Restate and expand the notes in IPDPS.txt.  We can show some API also.}

Requirements, design, and implementation issues are considered in this section.

\subsection{Requirements}
\textit{dMath} was created based on a number of key requirements:
\begin{itemize}
\setlength\itemsep{0em}
\item Support DNN pipelines efficiently with a variety of key algorithms;
\item Utilize multiple GPUs together with multiple MPI processes
to reduce time to solution for meaningful problem sizes for 
a range of commercial applications of interest to industry.
\item Exploit the latest GPUs from NVIDIA, latest InfiniBand from Mellanox,
and latest MPIs that utilize GDR in order to maximize inter-GPU performance.
\item Exploit the availability of PCI switching in order to gain density within
x86-64 servers;
\item Use modern C++ design and implementation strategies based on
object-oriented and meta-programming principles;
\item Allow users to write their algorithmic code without attention
to the details of how concurrency and data motion take place in the background
to effect scalable operation;
\item Support basic fault tolerance through checkpoint restart.
\end{itemize}
%\subsection{Overview}

\textit{dMath} runs as a set of MPI processes (subsets of {\tt MPI\_COMM\_WORLD}
less a master node), with a single master and many workers.  The
framework is initialized by the developer at the beginning of the
program.  At this point the workers enter run loops, waiting for
commands from the master node.  The developer uses \textit{dMath} like any
other mathematics library; the distributed computation is handled
internally, and implicitly, where the high-level user is unaware of
the distributed multi-GPU implementation.

\subsection{Encapsulation and Abstraction}
\textit{dMath} exploits many object-oriented design principles.  For example,
the Matrix class is an abstract class, defining the interface for a
Matrix.  Subclasses such as CudaMatrix then implement the virtual
methods.  This abstraction allows us to use the Abstract Factory
design pattern.  A MathFactory interface is defined, and concrete
subclasses of MathFactory are used to create specific varieties of
Matrix and Vector objects (CPU, CUDA, or Distributed).

\subsection{Managing persistent data in the GPUs}
The PCI switching architecture provides efficient access to multiple GPUs per root complex, but even with a single GPU per root complex, copying data across the CPU-GPU
boundary is undesirable.  \textit{dMath} data remains within the GPUs  (except when explicitly required by an algorithm to copy to the master process)
as it carries out the workflow required to implement a data parallel computation.  Caching is therefore a critical feature of \textit{dMath} needed to improve
the efficiency of CUDA-accelerated numerical computations.  The strong adherance to this model means that comparatively little of the resources of the multicore
CPUs are utilized.
% (could be value-engineered in certain dimensions in the future, so long as efficient PCI support is provided, and sufficient hardware
%threads are availale to drive CUDA Kernels in parallel).

\begin{figure}[]
  \centering
  \begin{tabular}{cc}
    \subfloat[]{%
      \includegraphics[width=0.2\columnwidth]{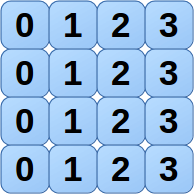}
    } &
    \subfloat[]{%
      \includegraphics[width=0.2\columnwidth]{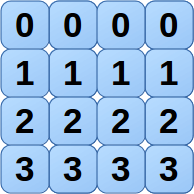}
    }%
    \\
    \subfloat[]{%
      \includegraphics[width=0.2\columnwidth]{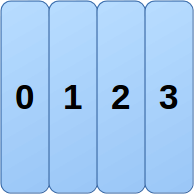}
    } &
    \subfloat[]{%
      \includegraphics[width=0.2\columnwidth]{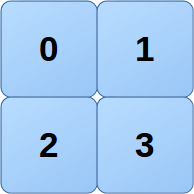}
    }%
  \end{tabular}
  \caption{Sample layouts for a distributed matrix.}
  \label{fig:data_layout}
\end{figure}

In \textit{dMath}, a distributed matrix is split into multiple non-overlapping
blocks, which are stored on individual workers.  Each worker is aware of
the layout of every matrix, allowing the workers to synchronize with
each other without the intervention of the master node.  The blocks
must be of the same size, with the exception of the last row or
column, which may be smaller.

The \textit{dMath} user is free to specify the layout of the blocks, taking advantage
of domain-specific knowledge.  Four sample layouts for a square matrix
are shown in Figure~\ref{fig:data_layout}.  The block size and worker
assignment must be chosen carefully.  Computation is most efficient
when the blocks are larger (GEMM for example), while pipelining is
easier when the blocks are smaller.

\begin{figure}[]
  \centering
  \includegraphics[width=0.7\columnwidth]{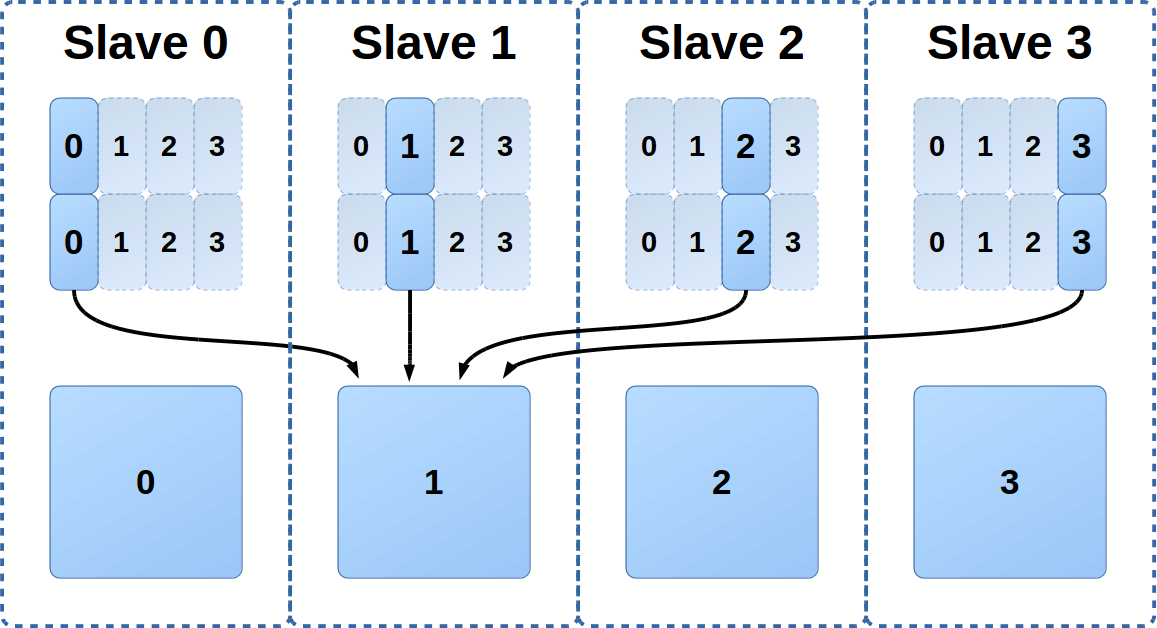}
  \caption{Replication of a distributed matrix.  The top row shows blocks stored by a worker in dark blue, and those stored by others in light blue.  The bottom row illustrates how each worker stores a replicated copy of the entire matrix.}
  \label{fig:replication}
\end{figure}

Often it is desirable to have a copy of a matrix on each worker.  In situations where the data rarely changes and memory is abundant, this sort of caching is beneficial.  \textit{dMath} supports this through the replication feature, shown in Figure~\ref{fig:replication}.  When the underlying matrix is updated, the workers automatically redistribute their data.

\begin{figure}[]
  \centering
  \begin{tabular}{cc}
    \subfloat[]{%
      \includegraphics[width=0.35\columnwidth]{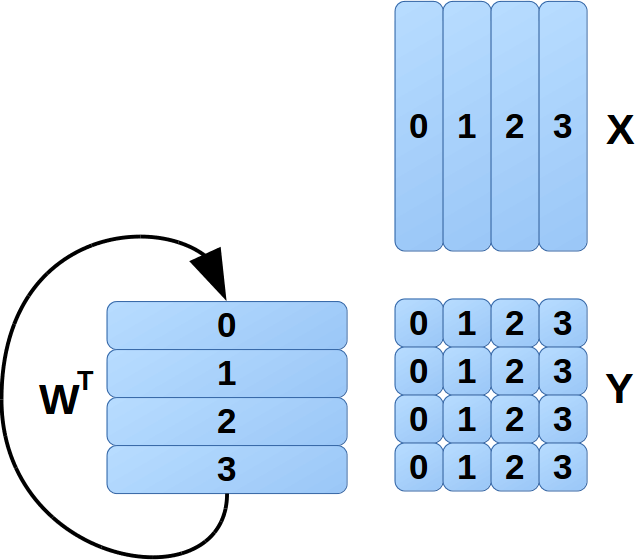}
    } &
    \subfloat[]{%
      \includegraphics[width=0.35\columnwidth]{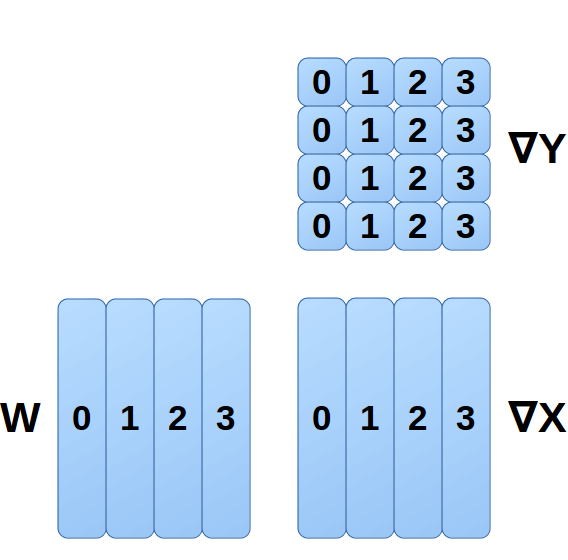}
    }%
  \end{tabular}

  \caption{ (a) Matrix multiplication for the forward pass through a fully connected layer ($Y = W^T X$).  The input data (top right) is already on the workers required for computing the output (bottom right).  The blocks of the weight matrix are cycled through the workers to perform the full computation (bottom left). (b) The backward pass ($\nabla X = W \nabla Y$) can be done without any communication between nodes, if the blocks of the weight matrix were cached on the forward pass. }
  \label{fig:cyclic_gemm_cache_fwd}
\end{figure}

Replication and caching allow us to efficiently perform backward passes when training the fully connected layer of a neural network.  As seen in Figure~\ref{fig:cyclic_gemm_cache_fwd}, the blocks of the weight matrix are rotated through the workers in the GEMM routine.  By caching these blocks in the forward pass, the backward pass can be done without any communication.

As the program runs, memory is often used and then discarded, both by the user and internally.  Instead of freeing this memory, \textit{dMath} pools such memory for later reuse.  This avoids the high cost of memory allocation in CUDA and the registration of this memory with the IB driver.

\subsection{Matrix Multiplication}

\begin{figure}[]
  \centering
  \includegraphics[width=0.5\columnwidth]{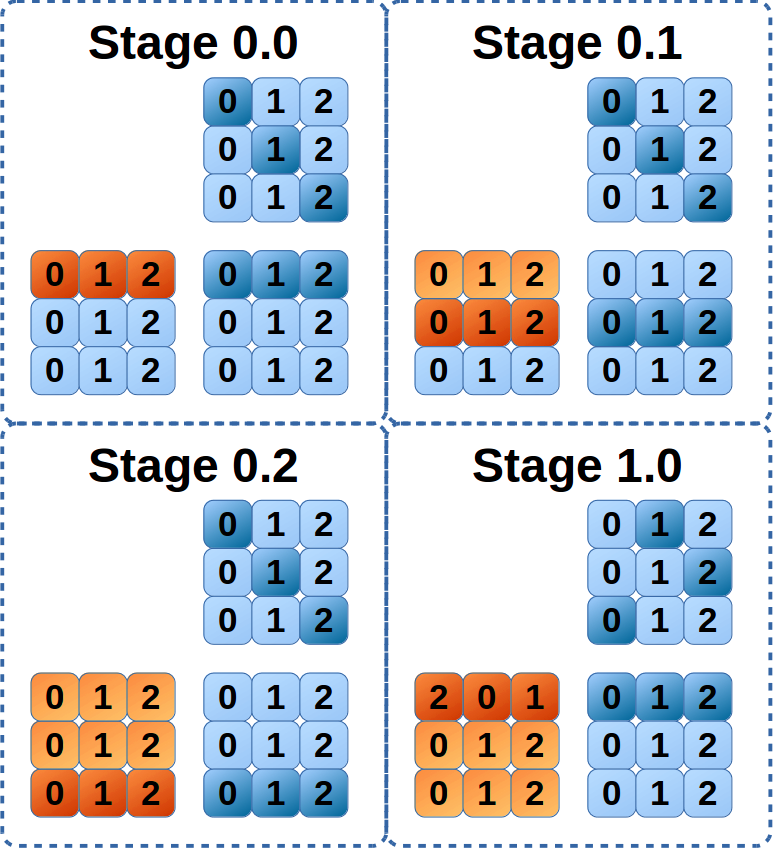}
  \caption{The first few stages of the matrix multiplication algorithm.  Dark blue indicates a block is being used in a GEMM call, dark orange indicates the block is being used in GEMM and that we're starting to cycle the block.  Light orange indicates that the cycling of block is still in progress.  The numbers indicate which worker the block was originally stored on.}
  \label{fig:cyclic_gemm}
\end{figure}

Matrix multiplication in \textit{dMath} is performed with a variant of Fox's algorithm~\cite{Fox198717}.  A one dimensional decomposition is used like in parts (a) (b) and (c) of Figure~\ref{fig:data_layout}.  The main difference from Fox's algorithm is that we have pipelined asynchronous rolling and no broadcasting.  The algorithm is divided into outer stages, based on the number workers used.  Each outer stage is divided into a number of inner stages, depending on how many blocks a worker's data is divided into.  On each step of the algorithm, every worker starts both an asynchronous GEMM call and an asynchronous cyclical transfer for the data in that row.  Each block has two buffers, used for sending and receiving.  On later stages, the algorithm will block until the required transfers and computation are completed before starting the next round.

A few steps of the algorithm can be seen in Figure~\ref{fig:cyclic_gemm}.  Stage $0.0$ initiates GEMM and transfers for the first row of $C$.  This is then repeated for the next two rows in stages $0.1$ and $0.2$.  On stage $1.0$, the algorithm waits until the computation and transfers are complete, ensuring that data is not overwritten while still being used.  Variants of this algorithm are implemented for transposes and rowmajor decompositions.

The architecture of the system (see Figure~\ref{fig:hw-architecture}) influenced our use of cyclic as opposed to broadcast based data transfer.  By maintaining one to one communication between devices, we can full PCIv3 speeds.  Communication tends to be the bottleneck in many scenarios, so this approach helps speed up the overall algorithm.

Other GEMM routines have been implemented in \textit{dMath} that can handle arbitrary matrix layout.  This generality comes at a price, however, and the cyclic variety presented above is preferred.

\subsection{Further Experience}
\subsubsection{Data loading}

While scaling up an application such as the training of a Deep Neural
Network (DNN), it's critical to load data fast enough so that the
algorithm doesn't stall.  \textit{dMath} handles this problem by storing
the datasets on high speed storage, including large amounts of host
memory or solid state devices.  While this can significantly reduce
loading time, distributing the data to the workers can also be a
significant bottleneck.  To remedy this issue, each worker loads its
subset of the data asynchronously in a thread.  The next batch is
typically ready by the time the current one is done processing. A strong argument towards in-memory storage is random access, important because it can give better sampling of the data, rather than using the same mini-batch periodically in every epoch of the data.

\subsubsection{Reproducibility}
The ability to reproduce results is incredibly important, and when
certain subroutines are stochastic in nature, one can get different
results that may be extremely difficult to duplicate. In \textit{dMath} we use
seed values that are distributed via the master node to workers to
ensure reproducible results in many cases. Nonetheless, there are
areas where concurrency and non-deterministic ordering of operations
can lead to small differences in results. For example, in the
distributed version of our \textit{AddRowColSumMatrix} subroutine we
sacrifice deterministic outcomes for speed and scalability, because
of summing in a non-deterministic way can produce different
results).
\subsubsection{Fault Tolerance}
The main goals of \textit{dMath} is to aid in tractability for computationally expensive machine learning techniques such as DNNs and for this reason fault tolerance is not one of driving forces as it adds too much overhead. That said, we rely on redundant arrays of inexpensive disks (RAID) 10 disk arrays, solid-state disks (SSDs) and most importantly a dynamic checkpoint variable. The checkpoint provides a means for the high-level user to grab a snapshot of the current system, in order to save the results to disk. We encourage such functionality because some experiments can run for hours, days, weeks, or even months and energy and development cycles are often expensive.

%%%%%%%%%%%%%%%%%%%%%%
%%%-------RELATED WORK 3 -----%%
%%%%%%%%%%%%%%%%%%%%%%
\section{Related Work}
%\label{sec:related-work}
\label{sec:related-work}

In the area of mathematical libraries, we note previous work by many
({\em e.g.}, \cite{Dongarra:1997:SUG:265932,Anderson:1990:LPL:110382.110385,VanZee:2015:BFR:2786970.2764454,dongarra-79}).  In
the area of GP-GPU-enabled libraries, there is significant work by
vendors and others \cite{DBLP:journals/corr/ChetlurWVCTCS14,nervana_neon,cufft,cublas,g80-paper,magma}.  In the
area of data-distribution-independent libraries, contributions from
several researchers are noted
\cite{van1990data,skjellum-toolbox,Skjellum95drivingissues,Bangalore95thedata-distribution-independent,conf/ppsc/AlpatovBEGMOGW97}.  Vast work on dense
matrix-matrix multiplication in parallel systems has been undertaken,
such as SUMMA \cite{summa}, PUMMA \cite{PUMMA}, and matrix
multiplication poly-algorithms \cite{DBLP:journals/concurrency/LiSF97}, among many
others.  Work on Strassen-based multiplication on CPUs and GPUs is of
potential relevance too ({\em e.g.}, \cite{DBLP:conf/hipc/LaiAES13}) in
problems that require low precision. 
There are two noteworthy libraries providing primitives for DL, cuDNN and neon~\cite{DBLP:journals/corr/ChetlurWVCTCS14,nervana_neon};
the former has broader and more generic implementation but both are intended for single-GPU use.

As mentioned above, in the area of high-performance interconnects,
Mellanox has devised the GDR-enabled optimization for Nvidia GPUs,
GPU-Direct RDMA (GDR) \cite{GDR-reference1}.  This, together with InfiniBand
networking provides the basis for COTS x86-64 or Power
\cite{ibm-power} clusters.  EDR InfiniBand \cite{EDR-IB}
represents the best approach to connecting such systems currently
available.  Numerous production and experimental clusters based on
InfiniBand, and GDR exist worldwide.  However, % in the area of
regarding PCI connectivity, we note the contribution of Cirrascale
\cite{cirrascale1}, which currently enables 96-lane PCI Gen3
connectivity on PCI root complexes in advanced x86-64 servers.  Adding
this component provides a design point that greatly improves density
and cost effectiveness, but also drives the need for efficient caching
and bandwidth utilization, which is noted above as one of {\em dMath}'s
strengths.

In the area of productivity for scalable parallel programming, PETSc
%\cite{petsc-web-page}
\cite{petsc-user-ref} is among the most of successful problem solving environments (PSEs)
for a domain-specific approach to productivity.  {\em dMath} follows a
similar approach of abstracting parallelism and providing a complete
set of primitives upon which to build an application, but in the
machine-learning, signal processing, and linear-algebra domain.
Matlab offers numerous toolboxes and parallel backends \cite{matlab} as well, 
focused on experimentation and prototyping and scientific exploration rather than for creating
production parallel software, or for handling multi-mode computations ({\em i.e.}, data parallel
and task parallel combinations such as {\em dMath} enables).

%There are a number of DL systems that must be credited as they are pioneering work in the field.
%Consider, the cloud-like computing systems from Google -- Dean~\textit{et al.}~\cite{40565} or Microsoft Research -- Chilimbi~\textit{et al.}  Project Adam~\cite{186212}, or more GPU-centric distributed frameworks from Stanford -- Coates~\textit{et al}~\cite{icml2013-coates13} or  Baidu -- Wu~\textit{et al.} Minwa system~\cite{DBLP:journals/corr/WuYSDS15}. Although these systems are interesting they are discerned to be single-pipeline, i.e. Automatic Speech Recognition, Image Recognition, or NLP, etc, based rather than true general purpose pipelines able to tackle any task. dMath was constructed based on base-primitive subroutines and this means we could easily construct AIR and ASR pipelines. This is incredibly useful as it allows various internal groups to utilize dMath for their specific problems and benefit from reduced experimental time and development cycles.

There are a number of DL systems that must be accredited as they are
pioneering work in the field.  Consider, the cloud-like computing
systems from Google -- Dean~\textit{et al.}~\cite{40565}, Microsoft
Research -- Chilimbi~\textit{et al.} Project Adam~\cite{186212}, and
more GPU-centric distributed frameworks from Stanford --
Coates~\textit{et al}~\cite{icml2013-coates13} and Baidu --
Wu~\textit{et al.} Minwa system~\cite{DBLP:journals/corr/WuYSDS15}. 
Although these
systems are interesting they are discerned to be single-pipeline,
that is, Automatic Speech Recognition, Image Recognition, or NLP, etc,
based rather than true general purpose pipelines able to tackle any
task. {\em dMath} was constructed based on base-primitive subroutines and
this means we could easily construct AIR and ASR pipelines. This is
incredibly useful as it allows various internal groups to utilize
{\em dMath} for their specific problems and benefit from reduced
experimental time and development cycles. Those libraries from
Microsoft \& Google, CNTK~\cite{CNTK} \&
TensorFlow~\cite{tensorflow2015-whitepaper} respectively, are the
closest to \textit{dMath}. CNTK was benchmarked but has stability
issues on several systems we evaluated and TensorFlow is incredibly
powerful but is not refined enough for performance at the time of this
submission. Lastly, MXNet is another candidate in the distributed
domain but lacks comparible scaling~\cite{MXNET_1}, given that MXNet
is slower than BVLC-Caffe, and we analyze Nvidia-Caffe, a faster
version of Caffe, we believe this is comprehensive enough review.

%{\em We may move this up--- here we will
%discuss how we are ``somewhat" like some libraries, but unlike 
%most... }

%%%%%%%%%%%%%%%%%%%%%%
%%%-------BODY SECTION 4-----%%
%%%%%%%%%%%%%%%%%%%%%%
\section{Algorithms and Degrees of Freedom}
\label{sec:Algorithms}
{\em dMath} combines a set of innovative ideas that simplify writing data parallel
algorithms while supporting key kernels.  We consider each briefly in turn.

\subsection{Algorithms}
{\em dMath} provides the following numerical kernels/operations distributed over MPI communicators comprising the workers.  Gather and scatter of these objects to/from the master is done when needed, but this is avoided except when absolutely essential.  Active objects are cached in GPU memory whenever possible, rather than moved in/out with each kernel invocation.  These
parallel operations are currenty supported:
%\begin{itemize}
%\setlength\itemsep{0em}
%\item a) dense matrix-vector and matrix-matrix multiplication (following Fox/BMR-type approaches currently \cite{Fox198717}) with
extensions; and, b)
%\item 
hundreds of algorithms and methods normative to a DNN computation pipeline.
%\end{itemize}
These building blocks allow a varierty of DNN-based pipelines to be created (implicitly  with parallel backends).

\subsection{Data Distribution Independence}
Algorithms in {\em dMath} are correct independently of the how the distributed objects are
mapped to the MPI processes (workers).  Unlike other popular data-parallel math libraries, {\em dMath} does
any needed communication to ensure compatibility, rather than limiting the distributions
to block-cyclic (and/or linear) 1D or 2D decompositions, such as used in Scalapack and other libraries.
Previous libraries ({\em e.g.,} \cite{van1990data,skjellum-toolbox})  achieve data
distribution independence and varying degrees of compatibility with rectangular matrices
and cartesian decompositions, but notably require that the objects be laid out compatibly at the beginning of the GEMM function,
rather than offering remapping services.  As with other libraries, the shape of
the data and concurrency changes the performance of {\em dMath} kernels (just not 
the correctness).  (It is notable that older libraries work hard to avoid any data reorganizations,
or leave that strictly to the user to implement, such as on top of {\tt MPI\_AlltoAll*}.)

\subsection{Precisions}
As with many object-oriented libraries, and even traditional API-based
libraries, multiple precisions are a key degree of freedom of {\em dMath}.
Nominally, single and double precision IEEE floating point operations
are provided in the parallel backend workers.  These are also the precisions
that are highly optimized in current high end GPU's, such as the Nvidia K80.
However, {\em dMath} also has added support for half-float precision, which is
available with functional support in CUDA 7.5, and has been supported in terms
of compressed storage in earlier CUDA versions too.  

Half-float is an IEEE-standard, 16-bit representation \cite{ieee-hfloat} that is suitable
for certain parts of computations, but may be totally unsuitable for
others where a longer significand is warranted.  While CUDA supports
half-float, current GPUs do not directly support high performance on
these datatypes with 16-bit arithmetic logic units (ALUs); such ALUs are anticipated in future GPU's, such as Nvidia's Pascal architecture.
At present, we utilize these operations with the understanding that
underlying CUDA BLAS will perform single-precision computations (even through the HGEMM interface),
rather than half-float computations.  Nonetheless, bandwidth and on-GPU storage savings
achieved by storing and moving 50\% less data across the CPU-GPU boundary are both of significant value.
At present, {\em dMath} is providing anticipatory support for future GPUs
where the 16-bit precision will actually run faster than float precision (32-bit). CUDA
HGEMMs directly support operations on this precision\footnote{Mixed precisions have
no cuBLAS API at present, but could follow the specifications in \cite{Li:2002:DIT:567806.567808}.}, but we also provide the option
to convert between half and float prior to using a CUDA BLAS call in order to optimize
 what come after this in the pipeline.

\subsection{Mixed Precision Computations}
From stage to stage, certain but not all stages of the data-parallel pipeline of interest
to {\em dMath} users can work effectively
varied precisions.  For this reason, {\em dMath} currently offers the ability to transfer
vectors and matrices at lower precision (with rounding), such as by moving a float matrix at half precision
and reconverting when needed prior to computation. This saves storage, memory \& network bandwidth.  As noted above, there are no fundamental
high performance BLAS that work on mixed precisions at present, so all the numerical objects must be
in the same precision before performing a CUDA BLAS kernel. {\em dMath} implements certain of its
own mixed precision operations at present, and apparently, with the advent of future GPUs that
support higher performance on half float, we may be able to justify the value of cuBLAS that
offer heterogeneous data types, rather than requiring pre- and post- conversions in a separate
kernel. This will be particularly relevant once next-generation GPUs provide for higher performance
with half precision.

Because of {\em dMath}'s object-oriented flexibility, experimenting with lower precision in certain stages of a pipeline
does not require a major rewrite of a code, thereby enabling the algorithm designer to explore the error implications
of changing precision and rounding.

%\subsection{Data-reorganization and data service model}
%{\em Notes:
%Optimization of computation that commonly occurs at the same time
%  Backward propagation through convolution layers, can put data, filter and bias gradient calculation in s\
%eparate streams.
%  Allows for higher GPU utilization.
%  Reduction can then be in parallel with other backward convolution computation.
%  Additional optimization can be done on the reduction operations by taking hardware and topology into acc\
%ount.
%}
%\subsubsection{Data caching, reorganization}
%\subsubsection{Mixed Precision data copies}
%{\em Note: this supports mixed precision workflows}

\subsection{Data Reorganization and Caching}
Where appropriate, {\em dMath} allows an algorithm to
\begin{enumerate}
\item reshape (including a change of concurrency and layout), over the same group of processes or a (super/sub)set
\item change precision during reshape.% ({\em e.g.,} float to double).%, float to half).
\end{enumerate}
This type of operation poses interesting challenges because such
operations combine CUDA kernels with MPI non-blocking point to point
and/or collective communications.  At present, because there is no way
to have kernel completion trigger MPI, or MPI completion trigger a
CUDA kernel, we have to have pre- and post- conversions operate
synchronously to the user thread before launching a non-blocking
collective.  In future, we may introduce Pthreads to support this, but
that introduces the potential for needing a multithreaded MPI
implementation with good multithreaded performance.  So efficient
reentrancy in MPI is becoming important, as is the need to be able to
have efficient merger of completion notification methodologies between
CUDA events, and {\tt MPI\_Wait/Test/Probe}\footnote{This gap motivates
  potential future standardization in the MPI Forum on
  interoperability between MPI and other parallel programming environments/notations like CUDA.}.

When the size of the objects is sufficiently small (as is
often the case in DNN problems), entire matrices fit in the individual
GPU memories (hence can be reused with reduced communication
and synchronization).  As such, our stepwise refinement that maximized the
concurrency of matrix formation need not correspond with the layout or
gross concurrency needed to minimize time to solution for a subsequent
step.  Remapping helps address the former issue. However, because of
the temporal locality of operations involved in DNNs, keeping segments
of matrices that arrive as temporary buffers during a matrix
multiplication can allow a subsequent matrix multiplication with the
same operand to function with communication suppressed.  Libraries of
which we are aware have not exploited such excess memory capacity; in
{\em dMath}, the data not only is retained, it is retained in the GPU memory
for maximum reuse performance.  We exploit the distinctive advantage
here of needing to solve problems for which architectural memory
capacity is sufficient for the GEMM phase to allow whole matrices to be
cached; we do not need to go to asymptotically large problems in order
to obtain peak performance for benchmarking purpose.  Other parts of
the computation utilize concurrency without replication.  This may be
counter intuitive to readers who always assume that any unscalabilty
is harmful.  By exploiting the provided resources fully, meaningful
problems are faster than without such unscalable replication;
furthermore, these problem sizes are not growing faster than the
available GPU memories for many use cases.

%The presence of the data replication service also allows for easy checkpoint restart support by user application, thereby simplifying fault tolerance for such problems.  The work can be periodically checkpointed if desired, and restarted if needed after a fault.  

%%%%%%%%%%%%%%%%%%%%%%
%%%-------BODY SECTION 5-----%%
%%%%%%%%%%%%%%%%%%%%%%
\section{Hardware Architecture \& Considerations}
\label{sec:HW_Arch_Considerations}
The dMath library provides for performance-portable parallel programs, as described in section~\ref{sec:dmath-architecture}.  It is important to indicate the underlying hybrid, heterogeneous architecture that Samsung Electronics has created in order to deliver the performance of cost-effective, high performance parallel algorithms to end users without extensive explicit parallel programming experience, with systems based integration of commodity off the shelf (COTS) technologies.  

\begin{figure}[!t]
\centering
\includegraphics[width=1.75in]{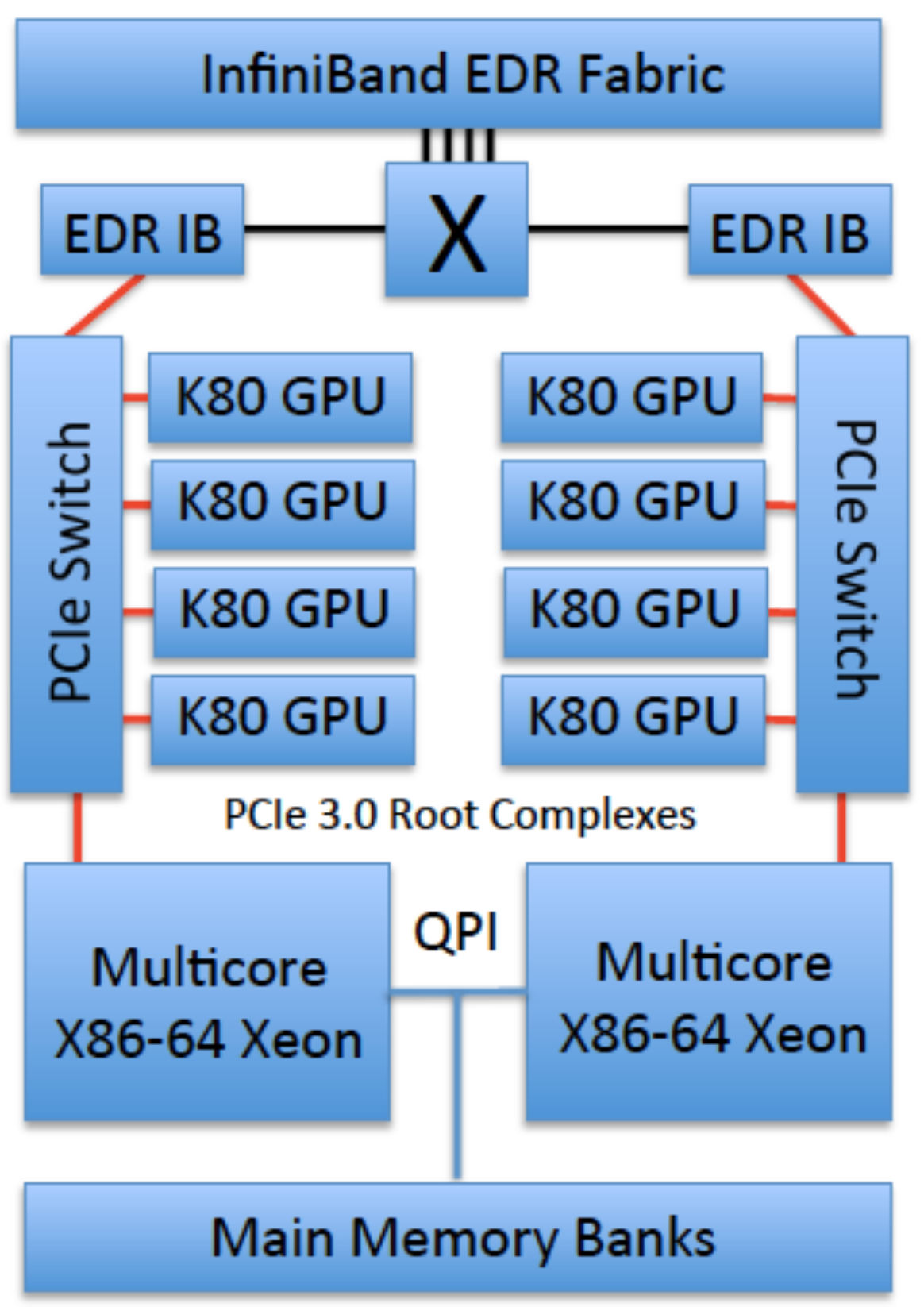}
\caption{Samsung Heterogeneous Multicomputer Node Architecture.}
\label{fig:hw-architecture}
\end{figure}

As illustrated in Figure~\ref{fig:hw-architecture}, the heterogeneous architecture utilized is comprised of X86-64 servers (multicore Haswell Xeon 2690 v2.0 \cite{IA-Haswell}) connected by dual EDR InfiniBand \cite{EDR-IB}, 512GB of host-memory, and SSDs for staging.  We utilize the dual PCI-Gen3 buses to support up to eight NVIDIA GP-GPU's and one InfiniBand EDR network adapter per scalable server unit root complex.  At present, we utilize NVIDIA K80 GPU's \cite{NVIDIA-K80} where auto boost is disabled and clocks set to 758Mhz, but this architecture is suitable for use with next-generation Maxwell \cite{NVIDIA-Maxwell} and Pascal GP-GPU's \cite{NVIDIA-Pascal} as well. When the text refers to a GPU it is that of a single-GPU not a dual GPU card, this is true for the experiments too.

The PCIe 3.0 crossbar switch fills an essential role, becaues it provides 
low-latency switching, and full bandwidth communication between any
pair. For instance, at any given moment, four GPUs can communicate and the host could be
communicating via IB at full PCIv3 speeds. As
long as one maintains pairwise communication between two devices,
communication does not degrade, and there is no congestion in this
scenario. Many of the subroutines within dMath are aware of the importance to use one to one communication to maintain optimal use of the h/w.  If we were to compare this COTS architectures having only
two GPUs per root-complex we can quickly see from
Table~\ref{table:comm_mediums_latency} \& \ref{table:comm_mediums_bw}
that the denser root complex scenario that utilizes CUDA P2P, and the
PCIe 3.0 crossbar switch, is favored over shared-memory, or Internode
IB EDR, for most use cases.

%{\em We need to emphasize the relatively low-end role of the
%Haswell cores... support GPUs, help with message
%passing, support staging of I/O, etc.}

%{\em Notes:  Compare storage of DB in SSD, RAID 10 Disk array through IB, Host Memory (NUMA issues)
%Loading of data from a single process is a large bottleneck.
%Ability to have each worker load their own data, allowing for high throughput
%By having loading asynchronous, image loading can be done while other work is going on, taking advantage o\
%f times with low CPU / PCI bus occupancy.
%RAM disk allows for random access when loading the data
%}

Modern computers have a variety of mediums for interprocess communication, e.g. shared-memory, CUDA P2P, InfiiniBand, etc, Figure~\ref{fig:hw-architecture} depicts these as blue, red, and black edges, respectively, and each has associated latencies and bandwidth. By understanding these mediums, and the workload, one can make effective use of them for specific tasks. As highlighted in Figure~\ref{fig:hw-architecture}, there are several means to communicate and Table~\ref{table:comm_mediums_latency} - \ref{table:comm_mediums_bw} show that communicating via shared-memory for GPU communication is very inefficient given high latency, and low bandwidth, whereas communication via Internode GDR is superior. Given the aforementioned observation we set out to introduce a new means for GPU intranode communication, where GPUs residing on different PCIe root complexes could communicate more efficiently. We worked with both OpenMPI and MVAPICH groups to provide the means to continue transferring host side communication via shared memory but used Infiniband for GPU to GPU traffic when GPUs were on different PCIe root complexes~\cite{OpenMPI2.0_pcc,mvapich2_pcc}. As one can see latency decreased and bandwidth increased but not to the level of Internode GDR; we are actively working to match the performance of both intranode and internode GDR traffic over Infiniband EDR. There is an important consideration that both versions of MPI supported the ability to disable shared memory but this would entail the need for small meta data, often transferred between processes, to traverse the IB stack and it is not as efficient as going through shared memory, e.g. consider latency for small transfers. The results reported are from an OpenMPI but MVAPICH2-GDR-PCC are similar.
In terms of real-world application performance, we will consider the use case of distributed Deep Learning (DL), where {\em dMath} v1.0 has a dedicated pipeline. The intradenode GDR showed training reductions in the order of 10-20\%, depending on the number of nodes and the number of fully-connected layers, where network communication is often the bottleneck and not computation. We quickly considered the situation of scaling beyond one compute node where contention to use the IB fabric for both intranode and internode communication could be an issue. However, testing revealed for the DL pipeline scaling to 2, 4, and beyond nodes, that is, 32-64+ GPUs, still benefitted from the intranode GDR but one should be mindful of possible congestion on the EDR fabric and adjust to use shared-memory, or hybrid, for intranode communication if the workload dictates it.

%note, only tested up to 4 nodes and 48 GPUs, we will soon have access to machines at cirrascale where we can test with 16-64 GPUs and we will have a chance to update this doc.
 
 We have shown and demonstrated alternative methodology for intranode
 communication that decreases latency and increased bandwidth;
 likewise scales well for Deep-Learning. The technique has shown to
 decrease runtime for GPU aware applications that rely on
 communication, such as training a CNN with fully-connected layers. The
 techniques are now available in both OpenMPI 2.0 and MVAPICH2-GDR-PCC
 from which all users of these MPIs can benefit.

\begin{table}
\caption{Communication Mediums: Latency} \label{table:comm_mediums_latency}
\begin{center}
\begin{tabular}{|l|l|l|l|l|l|l|}
\hline
\multicolumn{6}{|c|}{Latency ($\mu$s)} \\
\hline 
Transfer & Shared & Shared & IntraNode & InterNode & CUDA\\ 
Size     & Memory & Memory & IB - EDR  & IB - EDR  & P2P \\
Bytes    & Host   & GPU    & GDR       & GDR       &     \\ 
\hline 
\hline
0& 0.86 & \textbf{0.87} & 0.91 & 1.33 & 0.93\\
\hline 
1& 1.11 & 31.70 & 6.13 & \textbf{5.98} & 19.41\\
\hline
128& 1.28 & 26.25 & 5.83 & \textbf{5.77} & 15.51\\
\hline 
512& 1.54 & 26.20 & 12.00 & \textbf{11.58} & 15.33\\
\hline 
16384& 6.95 & 30.97 & 16.95 & \textbf{16.74} & 17.50\\
\hline 
524288& 138.61 & 163.39 & 218.72 & 157.12 & \textbf{80.91}\\
\hline
2097152& 501.10 & 515.71 & 458.22 & 425.37 & \textbf{279.04}\\
\hline 
4194304& 971.19 & 936.43 & 765.36 & 741.60 & \textbf{541.65}\\
\hline 
\end{tabular}
\end{center}
\end{table}

\begin{table}
\caption{Communication Mediums: Bandwidth} \label{table:comm_mediums_bw}
\begin{center}
\begin{tabular}{|l|l|l|l|l|l|l|}
\hline
\multicolumn{6}{|c|}{Bandwidth (MB/s)} \\
\hline 
Transfer & Shared & Shared & IntraNode & InterNode  & CUDA\\ 
Size     & Memory & Memory & IB - EDR  & IB - EDR   & P2P \\
Bytes    & Host   & GPU    & GDR       &  GDR       &     \\ 
\hline 
\hline
1& 1.76 & 0.06 & 0.58 & \textbf{0.68} & 0.13\\
\hline
128& 213.95 & 9.41 & 69.99 & \textbf{87.41} & 16.41\\
\hline 
512& 679.82 & 37.60 & 226.62 & \textbf{268.72} & 67.28\\
\hline 
16384& 5269.01 & 107.76 & 3558.15 & \textbf{3922.10} & 2336.16\\
\hline 
524288& 4540.58 & 4081.20 & 5298.05 & 6110.80 & \textbf{8984.97}\\
\hline
2097152& 4901.57 & 5148.11 & 7543.43 & 8105.62 & \textbf{9604.57}\\
\hline 
4194304& 5064.01 & 5266.48 & 7758.30 & 8657.90 & \textbf{9720.82}\\
\hline 
\end{tabular}
\end{center}
\end{table}

%%%%%%%%%%%%%%%%%%%%%%%
%%%-------EXPERIMENTAL RESULTS----%%
%%%%%%%%%%%%%%%%%%%%%%%
\section{Examples of performance and speed up}
\label{sec:examples}

\subsection{Deep Neural Network Training}

{\em dMath} provides hundreds of distributed, non-distributed GPU, and CPU
subroutines and benchmarking is a constant process via continuous
integration but annotating all those results is difficult; instead, we
will show generalized Matrix Matrix Multiplication (GEMM) and the
entire CNN pipeline for forward and backward propagation for raw
AlexNet~\cite{NIPS2012-4824} \& GoogLeNet v1. Profiling with AlexNet is beneficial as
it includes high ratio of Fully Connected (FC) layers, where GEMM is
often utilized, to convolutional layers. These FC layers are often
computationally light but heavy on communication and provide a good
worst-case scenario for our system. The most prominent models, as of
this publication, are those that use a higher ratio of convolutional
to FC layers such as VGG 16 and 19 layer models~\cite{Simonyan14c},
GoogLeNet~\cite{DBLP:journals/corr/SzegedyLJSRAEVR14},
BN-Inception~\cite{DBLP:conf/icml/2015}, inception-variants,
etc. These latter scale well when the batch size is increased, one has
to be cognizant of possible accuracy degradation with larger batches
but we have found many ways to stabilize the accuracy while scaling up
the batch size, e.g. batch normalization and PreLU have been able to narrow the gap between 256 \& 1024 batches to only 0.5\% for many models.  

\subsubsection{Weak Scaling DNN Tests}
We profile \textit{Expresso}, a Samsung internal forked version of Caffe powered by \textit{dMath} that employs hybrid parallelism~\cite{Krizhevsky14}, compared to a forked version of Caffe~\cite{DBLP:journals/corr/SzegedyLJSRAEVR14}, from Nvidia~\cite{nvidia_multi12}, which provides leading open-source multi-GPU scaling in a single machine via data - parallelism with models synced after every batch, i.e. nsync of one. We profiled both and achieved a 2615 FPS on 0.14 branch with cuDNNv4, {\em Expresso} obtaining 4198, FPS when utilizing AlexNet, 16 GPUs, and a 1,024 batch size. As one can see, the results favour \textit{Expresso} and one has to consider that because it is powered by \textit{dMath} one can scale past a single machine, as seen in the 32 and 64 GPU tests, whereas BVLC Caffe and Nvidia's own branch are limited to only intranode scaling. Likewise, the hybrid nature of the implementation means the models are spread across the devices and the total models size is limited to the aggregate device-memory of the total number of GPUs used during execution, versus data-parallel techniques, where the model must be able to fit into a single GPUs memory. This means, as models grow, \textit{Expresso} is a more favourable contender as it provides better intranode scaling, provides internode scaling, preserves accuracy, has the lowest memory footprint, and can support larger models. 

The experiments in Table~\ref{table:frameworkComparisons} use
a batch size of 1,024 for AlexNet 2-64 GPU, 512 for single, 1024 for
GoogLeNet 8-64 GPU and 128 for below 8 GPU. The single asterisk for Expresso designates non-optimal algorithm choice for convolutions due to memory constraints; hence super linear scaling from two to four GPUs. Whereas the double asterisk signifies the GPU device memory thrashing, i.e. need for costly alloc / dealloc is seen and impacts performance for nvidia-caffe 0.14.

\begin{table}
\caption{Framework Comparison (Weak Scaling)} \label{table:frameworkComparisons}
\begin{center}
\begin{tabular}{|c|c|c|c|c|}
\hline
Number  & \multicolumn{2}{c|}{AlexNet} & \multicolumn{2}{c|}{GoogLeNet v1}  \\
\multirow{2}{*}{of} &       \multicolumn{2}{c|}{1024 Batch}     & \multicolumn{2}{c|}{1024 Batch}\\
 &       \multicolumn{2}{c|}{ (FPS)}     & \multicolumn{2}{c|}{(FPS)}\\
GPUs &           Expresso v0.5 &    nv-caffe 0.14    & Expresso v0.5 & nv-caffe 0.14 \\
\hline
1        & \textbf{479}      &     413   & \textbf{115} & 102\\
\hline
2       &   *\textbf{940} &  **682 & \textbf{215} & 205 \\
4       &  \textbf{1996} &  1165 &  \textbf{370} & 341 \\
8       &  \textbf{3103} &  2204 &  \textbf{873} & **510 \\
16      & \textbf{4198} & 2615 &  1498 & \textbf{1515} \\
32      & \textbf{5187} & \textbf{-} & \textbf{2330}  & \textbf{-} \\
64      & \textbf{5786}  & \textbf{-} & \textbf{3025} & \textbf{-} \\
\hline
Memory  &   &  &  &   \\
GB      &  \textbf{2.29}  & 2.54  & \textbf{5.27} & 7.65 \\
(16 GPUs) & & & &\\
\hline
Accuracy   &   &  &  &   \\
(Top-1\%)    &   \textbf{55.38} &  55.14 & \textbf{65.39} & 64.96\\
\hline 
\end{tabular}
\end{center}
\end{table}

\subsubsection{Strong Scaling}
 \textit{Expresso} not only providing class leading performance for weak scaling it does also for strong scaling, as shown in Table~\ref{table:frameworkComparisons_strongScaling}. Strong scaling in this situation is when small batches are distributed and it is clearly seen that alternatives breakdown significantly. This is one of the fundamental goals of the underlining library \textit{dMath}, to provide the ability to experiment with extremely large models that are stored in device memory, without being constrained to only using subsets of GPUs because of poor strong scaling. Testing of CNTK was performed on both the regular \& 1-bit quantized SGD, whereas this feature was not available at the time of submission in \textit{dMath}. After extensive debugging we were not able to successfully run the multi-GPU version of CNTK for anything but a few iterations and cannot provide accuracy metrics, we hope to have complete results for CNTK for final revisions.

\begin{table}
\caption{Framework Comparison (Strong Scaling)} \label{table:frameworkComparisons_strongScaling}
\begin{center}
\begin{tabular}{|c|c|c|c|c|}
\hline
Number  & \multicolumn{4}{c|}{AlexNet}  \\
\multirow{2}{*}{of} &       \multicolumn{4}{c|}{256 Batch}\\
 &       \multicolumn{4}{c|}{ (FPS)} \\
GPUs &           Expresso v0.5 &    CNTK  & CNTK (1-bit) & nv-caffe  \\
           &                                   &   r2016-02-08 & r2016-02-08 & 0.14 \\
\hline
1        & 533      &     \textbf{580}   & 568 & 350\\
\hline
2       &   \textbf{915} &  487 & 485  & 711 \\
4       &  \textbf{1440} &  428 & 416 & 898 \\
8       &  \textbf{1702} &  - & -  & 970 \\
16      & \textbf{2008} & - &  - &  875\\
32      & \textbf{2104} & - & -  & - \\
64      & \textbf{2271}  & - & - & - \\
\hline
Accuracy   &   &  &  &   \\
(Top-1\%)    &   \textbf{58.59} &  - & - & 57.01\\
\hline 
\end{tabular}
\end{center}
\end{table}

\subsubsection{Accuracy}
Accuracy is an extremely important qualitative metric that is often overlooked in many distributed learning publications, we include those results and exhibit stabile accuracy when scaling GPUs. This is an important attribute of \textit{dMath} and it is because we solved the harder problem, that of hybrid parallelism, no matter the number of GPUs the accuracy is never impacted. The only impact on accuracy is that of batch size, nonetheless 1024 seems to be the sweet spot for good scaling and maintaining accuracy. We are not showcasing the top accuracy that can be obtained in Table~\ref{table:frameworkComparisons}-\ref{table:frameworkComparisons_strongScaling} for these models but simply identical hyper parameters, solvers, single crop, non-ensemble to demonstrate accuracy stability when scaling the number of GPUs. We would rather not speculate on why CNTK, or Nvidia Caffe, has lower accuracy but we can state that the \textit{Expresso} employs hybrid-parallelism and has provided stable accuracy as one scales the number of GPUs for a constant batch size.

\subsection{Matrix Multiplication}

We tested the performance of {\em dMath} on basic matrix multiplication, a fundamental of many computationally intensive problems. The results can be seen in Table~\ref{table:gemm_performance} where dMath provide leading intranode and internode scaling.  In some runs of dMath and cublas 7.5, there was insufficient memory to store the data (indicated by a dashed cell) in GPU device-memory whereas cublasXT can store in host memory.  For each of the listed sizes, a square matrix multiplication was performed 100 times.  The column for one GPU was done without {\em dMath}, using a simple cuBLAS 7.5 program.  For the remaining columns, the matrix was split up into equal parts in both rows and columns, with each row of blocks stored on a separate GPU device memory and maintaining the in-memory characteristics of the library. There are alternatives for intranode scaling like cublasXT or Magma~\cite{MR1484478}, the latter does not support GEMM, but these libraries can not scale past a single machine or achieve the same speedup.

For smaller matrix sizes, distributing to many GPUs was detrimental.  This was to be expected.  The performance of the cuBLAS GEMM routine is better with larger matrices as it utilizes the GPU more effectively.  Computation in matrix multiplication grows cubically, while storage / communication grows quadratically.  So at larger sizes, computation becomes the bottleneck and it becomes beneficial to distribute to more GPUs as shown in both dMath and cublasXT scaling.

\begin{table}[]
\centering
\caption{Matrix Multiplication Performance} \label{table:gemm_performance}
\resizebox{\columnwidth}{!}{%
\begin{tabular}{|r|rrrrrr}
\hline
\multicolumn{1}{|l|}{\multirow{3}{*}{Size}} & \multicolumn{6}{c|}{GPU times (s)}                                                                                                                                                                                                       \\ \cline{2-7} 
\multicolumn{1}{|l|}{}                      & \multicolumn{1}{c|}{1 GPU}           & \multicolumn{2}{c|}{2 GPUs}                                                 & \multicolumn{2}{c|}{8 GPUs}                                                 & \multicolumn{1}{c|}{32 GPUs}         \\ \cline{2-7} 
\multicolumn{1}{|l|}{}                      & \multicolumn{1}{l|}{cublas 7.5}          & \multicolumn{1}{l|}{dMath}          & \multicolumn{1}{l|}{cublasXt 7.5}         & \multicolumn{1}{l|}{dMath}           & \multicolumn{1}{l|}{cublasXt 7.5}        & \multicolumn{1}{l|}{dMath}           \\ \hline
4096                                        & \multicolumn{1}{r|}{0.052}  & \multicolumn{1}{r|}{0.035} & \multicolumn{1}{r|}{0.146}            & \multicolumn{1}{r|}{\textbf{0.021}}  & \multicolumn{1}{r|}{0.169}           & \multicolumn{1}{r|}{0.101}  \\
6144                                        & \multicolumn{1}{r|}{0.174}  & \multicolumn{1}{r|}{0.109} & \multicolumn{1}{r|}{0.589}            & \multicolumn{1}{r|}{\textbf{0.038}}  & \multicolumn{1}{r|}{0.259}           & \multicolumn{1}{r|}{0.104}  \\
8192                                        & \multicolumn{1}{r|}{0.413}  & \multicolumn{1}{r|}{0.245} & \multicolumn{1}{r|}{1.209}            & \multicolumn{1}{r|}{\textbf{0.076}}  & \multicolumn{1}{r|}{0.493}           & \multicolumn{1}{r|}{0.121}  \\
12288                                       & \multicolumn{1}{r|}{1.450}  & \multicolumn{1}{r|}{0.840} & \multicolumn{1}{r|}{3.268}            & \multicolumn{1}{r|}{0.433}  & \multicolumn{1}{r|}{1.420}           & \multicolumn{1}{r|}{\textbf{0.258}}  \\
16384                                       & \multicolumn{1}{r|}{3.340}  & \multicolumn{1}{r|}{2.034} & \multicolumn{1}{r|}{8.455}            & \multicolumn{1}{r|}{\textbf{0.528}}  & \multicolumn{1}{r|}{3.223}           & \multicolumn{1}{r|}{0.726}  \\
24576                                       & \multicolumn{1}{r|}{12.279} & \multicolumn{1}{r|}{6.809} & \multicolumn{1}{r|}{28.062}           & \multicolumn{1}{r|}{1.744}  & \multicolumn{1}{r|}{10.295}          & \multicolumn{1}{r|}{\textbf{1.091}}  \\
32768                                       & \multicolumn{1}{c|}{-}               & \multicolumn{1}{c|}{-}              & \multicolumn{1}{r|}{68.618}  & \multicolumn{1}{r|}{4.015}  & \multicolumn{1}{r|}{24.657}          & \multicolumn{1}{r|}{\textbf{2.389}}  \\
49152                                       & \multicolumn{1}{c|}{-}               & \multicolumn{1}{c|}{-}              & \multicolumn{1}{r|}{187.344} & \multicolumn{1}{r|}{14.016} & \multicolumn{1}{r|}{82.161}          & \multicolumn{1}{r|}{\textbf{5.260}}  \\
65536                                       & \multicolumn{1}{c|}{-}               & \multicolumn{1}{c|}{-}              & \multicolumn{1}{r|}{461.233} & \multicolumn{1}{c|}{-}               & \multicolumn{1}{r|}{192.380} & \multicolumn{1}{r|}{\textbf{10.592}} \\ \hline
\end{tabular}
}
\end{table}

\begin{comment}
\begin{figure}[!t]
\centering
\includegraphics[width=\columnwidth]{matrix_performance}
%\includegraphics[width=1.75in]{Figures/matrix_performance}
\caption{Matrix Multiplication Performance.}
\label{fig:matrix-performance}
\end{figure}
\end{comment}

\subsection{Experimental Results Conclusion}
We have shown an entire DNN pipeline that provides leading intranode / internode strong / weak scaling for a DNNs. We have also shown profiling of GEMM, a fundamental base-primitive for many domains, where we also exhibit leading performance. The hybrid nature of the system provides the ability to distributed the model across all available GPU nodes, whereas many others are constrained to the memory available in a single GPU, i.e. data-parallel techniques.

%%%%%%%%%%%%%%%%%%%%%%
%%%-------BODY SECTION 7-----%%
%%%%%%%%%%%%%%%%%%%%%%
\section{Heterogeneity and System Issues}
\label{sec:heterogeneity}
\label{sec:system-issues}
\subsection{Implications of Heterogeneity}
MPI implementations of which we are aware do not recognize the kinds and degrees of heterogeneity of systems
such as contained in our node architecture.  While almost all floating point computation is performed
in GPUs (and hence we don't have concerns about heterogeneity between CPU and GPU validity), the performance
of MPI collective operations and point-to-point operations is impacted by the memory hierarchy.  
For instance, data traversing QPI is substantially slower than data traversing EDR.  Furthermore,
transfers between GPUs in a single root complex outstrip transfer performance between complexes.
Typical MPI implementations make choices of collective operations based on a schedule, but with
a homogeneous view toward the performance of the underlying links and resources.  This is suboptimal
in our architecture.   This poses the need for topology awareness within MPI for the architecture,
and comprises near-future work for our project.

\subsection{Strong scaling and Excess Memory Capacity}
Whereas many problems derive from finite approximations of partial differential equations, and scale well with
added resources, our matrix-oriented problems are finite in dimension, and require strong scaling (Amdahl's law speedup) 
in order to be beneficial to {\em dMath} users.  Three basic stages of operations are involved in typical applications:
formation of matrices, data-parallel operations on those matrices ({\em e.g.,} comprising machine learning algorithms),
and potential post-processing analytics in memory (task parallel at the MPI process granularity).   Within the scope
of applications that we currently see as crucial to {\em dMath} users, large numbers of MPI processes (and underlying GPUs)
are suitable for matrix-formation/filling.  However, the finite size of the objects often means that less concurrency
at the MPI process level is needed to minimize time to solution, as noted above.  Furthermore, the problem sizes
encountered are small enough so that they often fit entirely in each GPU memory, not just distributed over a set of GPUs.
This ``over service'' of GPU memory at the linear algebra and data-parallel stage is a great benefit, because it means
that some operations gain additional performance through the elimination of communication needed in a purely distributed
environment.  We note such savings in Figures~\ref{fig:replication}, \ref{fig:cyclic_gemm_cache_fwd}.  In short, excess GPU
memory is of great value for relevant problem sizes.

\subsection{Maturity of the System Components}

The architecture and system created here composes a number of COTS architectures, several at the leading edge.  It is worthwhile
to note that the maturity of drivers, the PCI switch, servers, and GPUs has been for the most part quite good.  However, a number
of issues have naturally arisen on the ``bleeding edge'' involving the GDR-enabled MPI implementations, which are experimental.
We expect these implementations to continue to mature, and to expand support of GDR-enabled operations, including non-blocking
collective communication, which is not currently supported by either OpenMPI or MVAPICH2.  Topology awareness of blocking
and non-blocking collectives in MPI implementations that reflect the heterogeneous and hierarchical nature of the architecture will also be needed to
obtain efficient reductions and broadcasts as well.  Furthermore, new and better standardized MPI concepts that allow simultaneous and/or adaptive use of the
multiple data transfer paths available for certain transfers bears exploration; work considered in the MPI-4 Forum involving
persistence and channels may help support optimizations along these lines in future MPI versions.

%\section{Application Implications}
%Ease of use in programs that do analytics and need algorithmic speed up
%Maturity of the underlying COTS systems (MPI, hardware, drivers, ...)

%{\em Cam/Steven: What I'd like to do here is get a feel for the workflow between machine learning and analytics, and explain how dMath supports that at a high level, including ease of application coding, the persistence of the in-memory objects, and the ability of the app to go between MPI-style computations (backed by GPUs), and in-memory analytics... Not many people have, to my knowledge, near production systems that do both things with or without GP-GPUs to boost.  Lots of input on this section needed.

%Discussing the mixed precision modes - some 16, some 32 bit is really an important idea... including here OK, but it might be better farther up...
%}

%%%%%%%%%%%%%%%%%%%%%%
%%%-------FUTURE WORK-----%%
%%%%%%%%%%%%%%%%%%%%%%
\section{Future Work}
\label{sec:future-work}
%
% need's steven's blessing..
%
Significant near term opportunities present themselves for additions to dMath.  In the algorithmic area,
we are introducing additional variants for matrix-matrix multiplication, following \cite{DBLP:journals/concurrency/LiSF97}.  This strategy will allow dMath to use the fastest algorithm based on the specific shape and concurrency of the problem.

Additionally, we plan to utilize the Maxwell and Pascal architectures' high-speed, half-float precision.  Since certain algorithmic steps
can tolerate half-float precision, the introduction of half-float linear algebra will save memory bandwidth and transfer costs while providing adequate accuracy for certain numerical algorithms.

%In the area of hardware architecture, the following short-term evolution is expected:
%\begin{itemize}
%\item Upgrade from the 80-lane Cirrascale PCIe Switch Riser technology, to a future 96-lane version.
%This will be available by mid-2016.  This provides a 4:1 oversubscription model, since 16 lanes
%are devoted to the host, whereas 64 lanes are devoted to each of the five devices (four GPUs, and
%InfiniBand HCA).  This provides a signficant increase in available bandwidth per root complex.
%\item Utilize Pascal GP-GPU architecture when available in 2016.
%\end{itemize}
%Additionally, we will be scaling our existing system to a larger number of servers in order to support more simultaneous users (space sharing), and single larger-scale problems as well (``hero mode'').

Longer term, improvements in InfiniBand (EDR to HDR), represents an
additional opportunity, but this has to be keyed to greater overall
availability of root complex bandwidth.  Shorter term, we may explore
utilizing heterogeneous server notes, that support a higher degree of
bandwidth to GPU cards, in order to support certain stages of a
computation with fan-in, and limited floating point requirements.
Since dMath can remap efficiently, such a heterogeneous note would be
useful for early and late stages of some operations.  In particular, we
could support nodes with dual or quad HCA's per root complex, and dual GPUs,
vs. our standard of four GPUs and a single HCA per root complex.
However, before adding the level of heterogeneity to our system, we
will explore the achievable performance and total cost implications of
introducing a few of such nodes in a system.

%{\em Steven: you need to decide if we can discuss much here or not.  Like, we will use Pascals or not.}

%%%%%%%%%%%%%%%%%%%%%%
%%%-------CONCLUSION-----%%
%%%%%%%%%%%%%%%%%%%%%%
\section{Conclusion}
\label{sec:conclusion}
A new scalable parallel math library, {\em dMath}, was presented that provides a complete primitive pipeline for mathematics and has a
dedicated DL pipeline, we have shown experimental results, via \textit{Expresso}, that demonstrates superior scaling to all open-source frameworks.
When compared to previous work, we have
shown better single node scaling, and distributed multi-node
scaling, and both these results are important because they aid in tractability of
the problem while experimental time is reduced. We have shown class leading strong \& weak scaling that preserve accuracy when scaling up the number of GPUs. We believe the most important aspect is the
{\em dMath} framework provides a high-level programming language that implicitly
utilizes distributed multi-GPU programming and enables the end-user to
focus on their domain specific problem without the need to understand
parallel and distributed programming. We have demonstrated the effectiveness of \textit{dMath} via \textit{Expresso} but also have a version of of the popular open-source speech recognition library Kaldi~\cite{Povey_ASRU2011} powered by \textit{dMath} that shows the generality of the library. We have instrumented changes in OpenMPI and
MVAPICH2 MPI frameworks that have shown to decrease latency and
increase bandwidth. We shared how understanding of the hardware can lead to
better software, as in cyclic GEMM, a variant of Fox's algorithm. Last, we
have detailed further experiences that will inform all interested in
distributed learning or key mathematical operations at scale.

% or


%\appendices
%\section{Proof of the First Zonklar %Equation}

%%%%%%%%%%%%%%%%%%%%%%
%%%-------ACKNOWLEDGEMENTS-----%%
\section*{Acknowledgment}
The authors would like to thank all research collaboration partners
internal and external to Samsung Electronics Company. Specifically, 
we would like to acknowledge Rolf VandeVaart, from Nvidia, for his 
help with OpenMPI, Ohio State University MVAPICH group, Nvidia, 
Cirrascale, and Mellanox.

\ifCLASSOPTIONcaptionsoff
  \newpage
\fi

%\IEEEtriggeratref{46} %8


%\bibliographystyle{IEEEtran}
% test:
%\nocite{nervana_neon}
%
\nocite{Huss-Lederman93comparisonof}
\bibliographystyle{IEEEtran}
\bibliography{dmath-paper}

%\begin{thebibliography}{1}

%H.~Kopka and P.~W. Daly, \emph{A Guide to %\LaTeX}, 3rd~ed.\hskip 1em plus
%  0.5em minus 0.4em\relax Harlow, England: %ddison-Wesley, 1999.

%\end{thebibliography}

%\begin{biography}[{\includegraphics[width=1in,height=1.25in,clip,keepaspectratio]{mshell}}]{Michael Shell}

%\begin{IEEEbiography}[{\includegraphics[width=1in,height=1.25in,clip,keepaspectratio]{picture}}]{John Doe}

%\end{IEEEbiography}


%\vfill


\end{document}